# Anchor-based oversampling for imbalanced tabular data via contrastive and adversarial learning

Hadi Mohammadi [1], Ehsan Nazerfard, Mostafa Haghir Chehreghani

**Department of Computer Engineering**

Amirkabir University of Technology (Tehran Polytechnic), Tehran, Iran

{hadi_moh91, nazerfard, mostafa.chehreghani}@aut.ac.ir

**Abstract**
Imbalanced data represent a distribution with more frequencies of one class (majority) than the other (minority). This phenomenon occurs across various domains, such as security, medical care and human activity. In imbalanced learning, classification algorithms are typically inclined to classify the majority class accurately, resulting in artificially high accuracy rates. As a result, many minority samples are mistakenly labelled as majority-class instances, resulting in a bias that benefits the majority class. This study presents a framework based on boundary anchor samples to tackle the imbalance learning challenge. First, we select and use anchor samples to train a multilayer perceptron (MLP) classifier, which acts as a prior knowledge model and aids the adversarial and contrastive learning procedures. Then, we designed a novel deep generative model called Anchor Stabilized Conditional Generative Adversarial Network or Anch-SCGAN in short. Anch-SCGAN is supported with two generators for the minority and majority classes and a discriminator incorporating additional class-specific information from the pre-trained feature extractor MLP. In addition, we facilitate the generator's training procedure in two ways. First, we define a new generator loss function based on reprocessed anchor samples and contrastive learning. Second, we apply a scoring strategy to stabilize the adversarial training part in generators. We train Anch-SCGAN and further finetune it with anchor samples to improve the precision of the generated samples. Our experiments on 16 real-world imbalanced datasets illustrate that Anch-SCGAN outperforms the renowned methods in imbalanced learning.

## 1 Introduction

Data access and storage have become more efficient with the advancement of hardware and internet technologies. Extracting information from raw data is vital for various individuals and businesses. Therefore, various machine learning models have been developed for supervised and unsupervised learning. However, despite the constant improvement of classification models and the emergence of new methods over time, imbalanced datasets remain a challenge for such models. Imbalanced data are characterized by a distribution in which the majority class outnumbers the minority class, resulting in a skewed representation of instances. Many real-world practices are associated with imbalanced datasets. For instance, in bank transaction fraud detection [1], a large portion of transactions are regular and fraudulent transactions are scarce. In medicine and healthcare [2] severe medical conditions among patients are practically rare. Other examples appear in customer classification [3], fault detection [4], software defect detection [5] and text classification [6]. Classification models typically presuppose a balanced class distribution during training [7]. Consequently, these models may encounter difficulties when learning the distribution of minority classes, resulting in a potential bias towards the majority class distribution. Addressing imbalanced data involves enhancing classification performance to more accurately represent the minority class's distribution, a challenge commonly referred to as the imbalance learning problem.

---

[1] Corresponding author
   E-mail address: hadi_moh91@aut.ac.ir

Multiple approaches have been suggested to solve the problem of imbalanced learning, which primarily fall into two main groups: data-based and algorithm-based strategies. Algorithm-level approaches use newly designed or modified algorithms to deal with imbalanced learning tasks. Common techniques include cost-sensitive methods [8], which reweight minority samples and adopt an adjective function to label the minority class more accurately, and ensemble learning methods [9] which aggregate the decision of multiple classifiers to decide on the outcome of classification. Cost-sensitive methods are controlled by specific domains, and the model ability is compromised with changes in the data field. Ensemble methods are mostly combined with other approaches to enhance classification results [10].

Data-level strategies employ sampling techniques to achieve dataset balance by: i) oversampling the minority class [11], ii) undersampling the majority class [12], or iii) applying hybrid sampling that integrates both methods [13]. It is worth noting that undersampling techniques are prone to information loss due to the distortion of majority class distribution [14]. Oversampling techniques are more robust to data distribution changes than undersampling methods.

Oversampling methods are primarily divided into classical oversampling methods and deep generative models. Classical methods are primarily concerned with regional information to generate new instances. Popular classical methods include random oversampling (ROS), which duplicates randomly selected minority instances, and synthetic minority oversampling (SMOTE) [15], which utilizes the nearest neighborhood approach along with linear interpolation between two minority samples to generate new samples. Deep generative models, on the other hand, are based on neural networks such as Generative Adversarial Network (GAN) [16], Variational Autoencoder [17] or Normalizing Flows [18] to generate synthetic samples. Classical methods are popular due to their simple design, low computing cost, and impressive performance [19]. However, considering their inherent randomness in selecting minority samples, consistent classification outcomes are not guaranteed. SMOTE disregard overlapping area between two classes. SMOTE-based methods such as B-SMOTE [20] and ADASYN [21] often create minority class instances in locations primarily dominated by the majority class, leading to the production of noisy samples [22].

The generative adversarial network has emerged as a class of generative models trained to mimic datasets' original distribution and directly synthesize artificial data. Due to the superior performance of GAN in generating realistic data, it has been expanded in image, text, video, and voice generation. GAN-based schemes have also been utilized as an oversampling tool to overcome the class imbalance problem. The Conditional Generative Adversarial Network (CGAN) [23] is an enhanced edition of the GAN that can learn the distinct distributions of various classes by incorporating supplementary information into the GAN framework. In this scenario, CGAN can be adeptly trained to capture the distributions of classes from minority and majority samples. In contrast, a standard GAN is typically limited to learning only one class distribution. This distinction can improve the reliability of the synthesized minority class samples, contributing to more representative samples for the final classification task.

This study presents an oversampling scheme leveraging anchor samples in the influential borderline regions. Several scholars have advocated utilizing data proximal to the decision boundary, showcasing its efficacy in various applications [24, 25]. Likewise, we select samples from minority and majority classes in proximity to the decision region while excluding hard samples from both classes. This creates a smooth decision boundary with fewer overlapping samples. In this regard, a multilayer perceptron (MLP) classifier is trained on anchor samples to gain prior knowledge about the borderline region. Further, a novel variation of CGAN is designed that is complemented by a previously trained MLP classifier and includes two minority and majority class generators. The prior network serves as a class representation for the discriminator and assists generators in contrastive and adversarial learning tasks. Finally, after training the generative network, we use anchor samples to finetune and improve model capability in generating realistic samples.

Our main contributions are as follows:

- Firstly, Anch-SCGAN occupies a multiphase anchor selection preprocessing scheme that provides a more streamlined and generalized view of the decision boundary. An MLP classifier is trained on anchor samples to provide elementary knowledge of the borderline region.
- Secondly, we employ the pre-trained MLP classifier to supply the generator and discriminator with supplementary information and relieve the scarcity of minority samples during training, which enhances the model's capability to generate near-real samples.
- Thirdly, we present a new generative model based on adversarial and contrastive learning consisting of a novel network architecture and a customized loss function. We design an adapted CGAN that utilizes two generators for minority and majority classes and a discriminator supported with additional class representation input. Further, we integrate a contrastive loss function with reformed anchor samples to create an anchor loss function and add it to the primary loss functions for generators.
- Fourthly, we design a procedure to stabilize the adversarial training of generators in GAN by introducing a score function derived from the prior MLP network. This strategy will help the generators to preserve their performance throughout the training. We provide detailed experiments on 16 imbalanced datasets. Our experimental results show that our proposed model outperforms well-known existing oversampling techniques in addressing the imbalanced learning problem.

The rest of this paper is structured in the following manner. section 2, delivers an extensive discussion of the existing literature on imbalanced learning. In Section 3, we provided a comprehensive explanation of our suggested approach. In Section 4, we report the outcomes of our experiments and demonstrate the practicality of our proposed strategy. Finally, the paper is concluded in Section 5.

## 2 Related work

In this section, we review three main approaches for addressing the imbalanced learning problem: data-level techniques, algorithm-level techniques and deep generative methods. Given that in this paper our primary focus is on generative models, we delve into these methods in a separate section.

### 2.1 Data-based methods

As mentioned in the previous section, data-level methods involve sampling strategies to mitigate the imbalance between minority and majority class. These methods encompass a range of techniques such as oversampling, undersampling, and hybrid sampling.

Oversampling goal is to increase the size of minority instances to adjust the distribution between minority and majority class. Chawla et al. [15] proposed SMOTE as an oversampling scheme which generate new sample through linear connection between two randomly selected neighbors. Multiple studies investigate the production of synthetic samples at the boundary location, highlighting the significance of boundary data in determining the final decision boundary. Han et al. [20] proposed Borderline-SMOTE where only minority sample from boundary region is selected for generating new samples. The selection process of candidate sample is based on the ratio of minority and majority samples among its surrounding samples. Haibo et al. [21] presented ADASYN, which uses an idea similar to the idea of Borderline-SMOTE. ADASYN assigns a value for every minority instance. The value is defined by the proportion of minority instances in the neighborhood. samples with higher values have higher chance to be chosen as candidate for generating new sample. Inspired by SMOTE, Yang et al. [26] proposed an oversampling method called SD-KMSMOTE which pay attention to majority class distribution surrounding minority class samples. In this method minority samples are clustered via K-mean procedure. Afterward, minority class instances that are isolated or distant from clusters are discarded from the dataset. Eventually, different values are assigned to each cluster, and clusters with higher values contribute to higher numbers of generated synthesis samples. Xie et al. [27] presented Gaussian Distribution based oversampling (GDO) as a solution for an imbalanced learning problem. GDO employs a probabilistic approach to choose base instances from the minority class for oversampling purposes, considering the density and distance information presented by the minority and majority distribution. Unlike creating new artificial instances in a linear manner, GDO employs a Gaussian distribution to produce synthethic samples.

Undersampling methods equalize the class distributions by deducting the instances in the majority class while maintaining the overall data distribution. Several studies aim at the problem of class overlap in imbalanced data. For example, Vuttipittayamongkol et al. [28] introduced four neighborhood-based undersampling methods which focus on overlapping area and eliminating majority class in such area to increase the visibility of minority samples. Guzmán-Ponce et al. [29] proposed two stage undersampling method called DSIG-US to alleviate class imbalance. First DBSCAN clustering procedure is utilized to identify and discard noisy instances from majority class. Then a graph-based algorithm is used to reduce the proportions of instances belonging to the majority class until a balanced dataset is achieved. Koziarski et al. [30] proposed radial-based undersampling (RBU), to overcome the limitation of neighborhood-based approaches. RBU utilize potential function to calculate mutual information of every majority sample and discard any with highest mutual information.

Hybrid methods combine oversampling and undersampling schemes to enhance the performance of the sampling process. Batista et al. [31] introduced a hybrid sampling technique that first apply SMOTE for oversampling minority class and then use data cleaning methods to create proper class clusters. Further, [32] introduce SMOTE-WENN which combines SMOTE with novel data cleaning method called weighted edited nearest neighbor rule (WENN). WENN discovers and removes unsafe minority and majority class samples using modified distance function along with k-NN. Finally, Zhu et al. [33] introduced the Evolutionary Hybrid Sampling method (EHSO). EHSO improves the decision boundary by eliminating unnecessary samples in the majority class and making it clearer. Then, random oversampling is employed to adjust the distribution of both the minority and majority classes.

In data-based methods, oversampling methods often struggle to produce samples that resemble the original distribution as new instances are generated mainly based on closeness to their minority neighbors. In addition, minority samples generated near the decision boundary often fall in the region of the majority class, increasing the risk of overfitting on noisy samples. Undersampling methods increase the risk of removing valuable information due to the elimination of the majority of samples. Hybrid methods are derived from both oversampling and undersampling methods. So, they share the same defects. This study aims to overcome the limitations of conventional approaches by acquiring knowledge about the underlying data distribution and generating samples from a global perspective.

## 2.2 Algorithm-based methods

Algorithm-based methods are designed to adjust existing algorithms to mitigate the effect of class imbalance and improve the classification accuracy of the minority class without compromising the overall accuracy. Cost-sensitive methods are among the most popular algorithm-based approaches for addressing the imbalanced learning problem. These models assign high penalties to errors involving the minority class, encouraging the classifier to give more importance to correctly classifying these instances. For example, Zheng et al. [34] introduced cost sensitive hierarchical classification for imbalanced data called CSHCIC. The first step involves using a hierarchical tree structure to break down a big classification problem into smaller subclassification tasks. Following that, a cost-sensitive method is individually utilized in each subclassification task. Wang et al. [35] proposed Adaptive FH-SVM which reconstructs hinge loss by integrating the scaling factor from focal loss, referred to as FH loss. In addition, an enhanced model along with modified FH loss is designed to solve the inconsistency between SVM and FH loss.

Ensemble methods can also be effective in handling class imbalance by combining multiple base learners trained on different subset of data. For instance, Wang et al. [36] developed a technique to improve the AdaBoost algorithm for classifying imbalanced data. They achieved this by adjusting the weighted vote parameters based on the overall error rate and the minority class's accuracy rate. Several studies combine ensemble learning with cost-sensitive learning. For example, Tao et al. [37] proposed a self-adaptive cost weights-based SVM cost-sensitive ensemble for classifying imbalanced data. This approach recruit cost-sensitive Support Vector Machines (SVMs) as the base learners and adapts the typical boosting algorithm to a cost-sensitive setting.

Although algorithm-based methods are widely used in imbalanced learning, their applicability can vary across different fields. Besides, cost-sensitive approaches require assigning different costs to misclassified samples, which is challenging and demands proficient knowledge. Ensemble methods' performance is hindered by the scarcity of minority classes; thus, they are often combined with data-level or cost-sensitive methods. The study introduces a data-driven oversampling strategy independent of the classification algorithm and suitable for testing across various disciplines.

## 2.3 GAN and Deep Generative based methods

Generative Adversarial Networks (GANs) have demonstrated their effectiveness in capturing data distributions and are recognized as a significant method for data augmentation. Consequently, various studies have recommended the use of GANs to address imbalanced learning problems, where they can generate synthetic instances to balance the dataset [38, 39]. Unlike traditional oversampling methods, GANs have the ability to recognize the underlying distribution of original data from global point of view and generate artificial data by using intricate transformations. Ahsan et al. [40] combines Borderline-SMOTE and generative adversarial network to generate synthetic minority samples. In this method, instead of gaussian noise, generated samples from Borderline-SMOTE is passed to the generator and training of GAN is performed under this modification. Joloudari et al. [41] study the performance of Conditional Generative Adversarial Network (CGAN) on image dataset with different imbalance ratio, class overlap, data dimension and sample size. This study demonstrates the effectiveness of CGAN as synthetic sample generator under multiple difficulty factors. Son et al. [24] proposed an oversampling scheme that defines borderline class data near the majority class and trains a CGAN on the majority, minority, and boundary classes. However, CGANs are known to be prone to mode collapse and can often suffer from unstable training [42]. One strategy for addressing mode collapse is modifying the objective function. Jo et al. [25] introduced a technique known as minority oversampling along the boundary with a generative adversarial network (OBGAN). OBGAN utilizes two distinct discriminators, one for the minority class and another for the majority class, alongside a generator. The generator is adept at modeling the data distribution of both classes, with an emphasis on the regions near the class boundary. In addition, the framework and loss function of OBGAN are specifically intended to minimize the chance of mode collapse, ensuring a diverse and representative generation of samples. Ding et al. [43] developed a hybrid method named RGAN-EL, which combines generative adversarial networks with ensemble learning. This framework incorporates a sample selection technique based on the roulette wheel selection method to improve the emphasis of GANs on the overlapping region between classes. RGAN-EL consists of two additional loss functions for the generator, as well as a final step for filtering noise samples, with the goal of improving the quality of the generated samples. Ding et al. [44] designed the RVGAN-TL model, which integrates Generative Adversarial Networks (GANs) with transfer learning to address the issue of imbalanced learning problem. In order to enhance the performance of the GAN, a similarity loss function is incorporated into the generator. In addition, a Variational Autoencoder (VAE) is employed to produce a latent vector which serves as an input for the generator in GAN. Schultz et al. [45] present Convex-space Generative Network (ConvGeN), a deep generative model that integrates convex space learning with a deep generative framework. Unlike typical GAN models, in ConvGeN, two network architectures—a generator and a discriminator—collaborate during training. The discriminator trains to distinguish between synthetic samples and a shuffled batch of majority class samples, while the generator learns to determine the coefficients for convex combinations of minority class samples.

While numerous studies have utilized GANs to deal with the difficulties of imbalanced learning, these models still possess intrinsic limitations. The shape of the decision boundary is affected by the existence of minority and majority samples in regions where classes overlap. Therefore, CGAN and its variants may be biased toward learning the distribution of the majority class and neglect the minority class data. This results from an abundance of majority samples in the overlapping area. Training GANs only on minority class distribution and ignoring the presence of the majority class lead to over-generalization. Without constraints, produced samples will be pushed to the majority class region. Based on this, we design an anchor selection strategy focusing on samples from both classes close to the decision boundary. After the initial selection, a classifier is trained on a balanced anchor dataset to establish a fair decision boundary and hinder the effects of the dominant class in this region. This classifier contributes valuable information to the proposed deep generative network. Additionally, we integrated contrastive learning-based anchor loss into the minority and majority class generators to constrain the generators to produce samples that closely resemble the same class distribution.

Table 1, provides a concise explanation of the contributions made in recent literature. The appearance of the "DB" column in the table indicates whether the proposed method takes into account the influence of the decision boundary or not.

**Table 1**
Summary of recent works.

| Group | Subgroup | Literature | Methodology | DB |
|---|---|---|---|---|
| Data-based | Oversampling | [15] | SMOTE (KNN + linear combination) | ✗ |
| | | [20, 21] | SMOTE + KNN | ✓ |
| | | [26] | SMOTE + clustering | ✓ |
| | | [27] | Gaussian distribution Oversampling | ✓ |
| | GAN and Deep generative models | [38, 41] | Conditional GAN | ✗ |
| | | [40] | GAN + Borderline SMOTE | ✓ |
| | | [24] | Conditional GAN + Borderline class | ✓ |
| | | [25] | Conditional GAN + Modification function | ✓ |
| | | [43] | Modification GAN + Ensemble learning | ✓ |
| | | [44] | Modification GAN + Transfer learning | ✓ |
| | | [45] | Convex learning + Deep generative model | ✓ |
| | Undersampling | [28] | KNN + Neighborhood search | ✓ |
| | | [29] | DBSCAN + graph-based algorithm | ✓ |
| | | [30] | Radial based function | ✓ |
| | Hybrid methods | [31, 32] | SMOTE + Data cleaning | ✗ |
| | | [33] | Evolutionary Undersampling + ROS | ✓ |
| Algorithm-based | Cost-sensitive | [34] | hierarchical classification + Cost-sensitive | ✗ |
| | | [35] | SVM + modified FH loss | ✓ |
| | Ensemble methods | [36] | Improved AdaBoost | ✗ |
| | | [37] | SVM + Cost-sensitive + Ensemble learning | ✓ |

## 3 Methodology

At this point, we provide an in-depth explanation of our approach. The overview of our algorithm for addressing class imbalance is shown in Fig. 1. The figure illustrates the initial step of dividing the dataset into two sets: the training and the test data. We solely rely on the training set to construct a balanced dataset and obtain a better training performance. This process includes: 1) The borderline sample selection is carried out on the training set to gather anchor samples, 2) anchor samples are used to train the MLP classifier, 3) The classifier is utilized to filter complex and noisy samples from minority and majority class data that facilitate the generalization of the model, 4) the more robust dataset is then used to train our proposed model, 5) An additional training phase is conducted with borderline anchor samples to slightly adjust the weights of the network and enhance the effectiveness of pseudo sample generation. 6) The ultimate minority class generator synthesizes minority samples to achieve an even distribution of data classes. 7) Finally, the classification model is developed on the balanced train set and assessed on the test set.

### 3.1 Anchor sample selection

In general, instances near the decision boundary play an active role in specifying the outcome of classifying results. In an imbalanced dataset, minority samples in the boundary area are usually sparse, resulting in a skewed data distribution favoring the majority class. The priority of CGAN or any generative model, as an oversampling scheme in imbalanced learning, is to replicate the minority sample data and, in particular, to understand the distribution of the minority class in the boundary region to generate better samples. One approach is to neglect majority samples and train the GAN solely on minority samples, focusing on borderline instances. However, this proposal might diminish the integrity of the generated samples and ignore the problem rather than deal with it. Hence, it is more efficient to train the entire training set with the presence of two class distributions. In this scenario, the generated minority samples are less likely to be formed in the majority class distribution area, improving the oversampling scheme's viability. Based on what has been discussed, we introduce a sample selection method which consists of three parts: 1) minority sample selection, 2) majority sample selection, and 3) establishing the final balanced anchor samples. The main goal is to select the most crucial instances from both classes to provide them with more learning opportunities. The overall scheme of our anchor sample selection is presented in Algorithm 1.

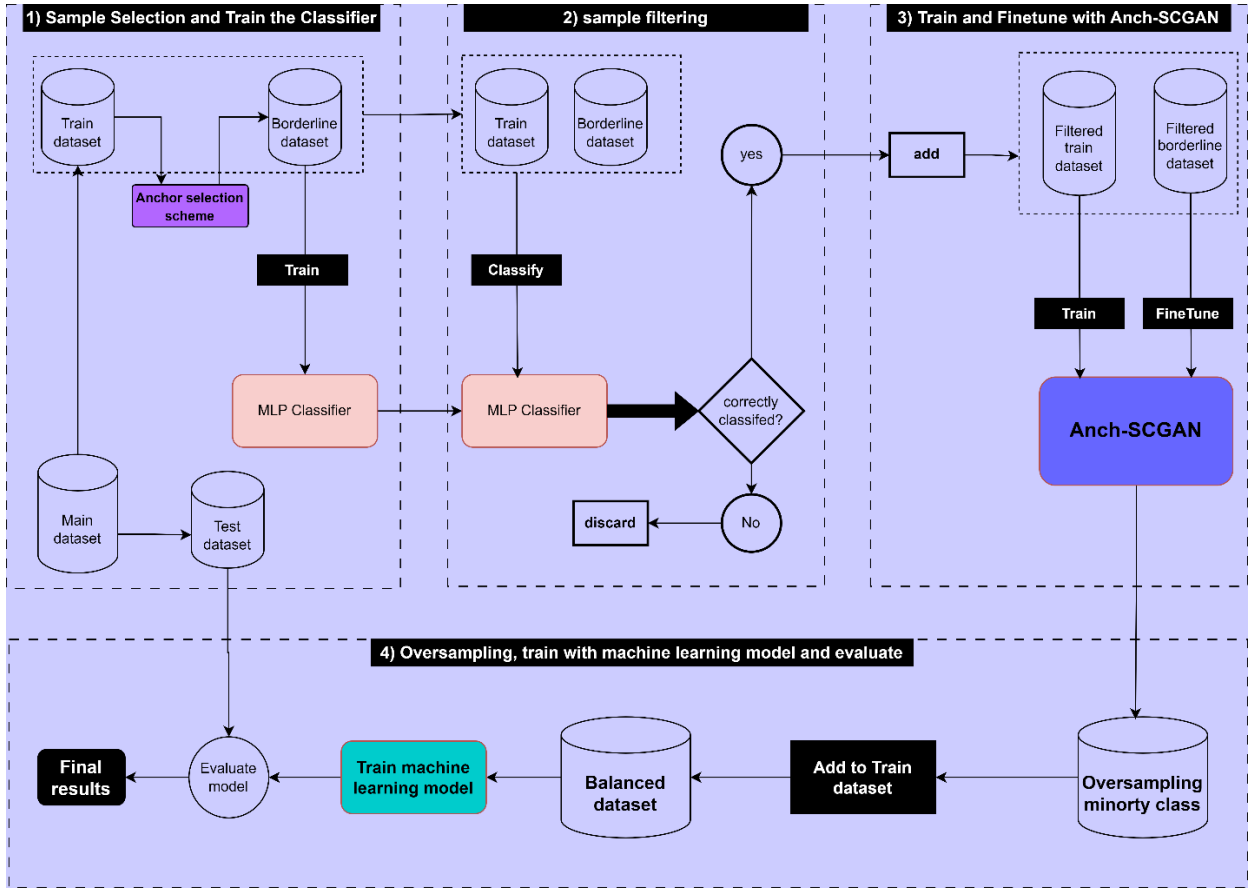

**Fig. 1.** graphical scheme of our proposed method for imbalanced learning.

### 3.1.1 Minority sample selection

Considering the importance of minority samples, we can select instances in proximity to majority-class samples. We indicate the $k$ closest instances to each instance $x$ in the minority class and denote them as NN($x$). The sample $x$ is regarded as an anchor sample if at least one NN($x$) sample belongs to the majority class. However, if all NN($x$) samples belong to the majority class, $x$ can be discarded from the data as a noise sample depending on the imbalance ratio intensity. Noise samples can cause problems for learning algorithms. In addition, a second group of minority samples will be selected if they have an adjacency relationship with borderline minority samples. The remaining minority samples, distant from the boundary area, have a diminished impact on defining the decision boundary and may or may not be included as anchor samples depending on the prominence of the opposite class sample.

### 3.1.2 Majority sample selection

Prior to selecting the majority anchor samples, it is necessary to discard any majority instances in the overlapped area. Removing majority samples near minority samples boosts the detectability of minority instances in the boundary region. As a result, the learning algorithm can learn the minority class distribution more effectively. Every majority instance is discarded if it exists within the neighbors of more than $k/2$ minority instances. If any majority instance is located within the neighbors of less than $k/2$ minority instances but at least one, it is regarded as an anchor majority sample.

---

**Algorithm 1** anchor sample selection.

---

**Input:** positive samples $P$ ; negative class samples $N$ ; nearest neighbor parameter $k$.
**Output:** selected anchor set $T_{anch}$
1: $T_{min} = \{\}$ ; $T_{maj} = \{\}$;
*// minority sample selection*
2: **for** each $x \in P$ **do**
3:      $NN(x)$: $k$ closest neighbor of x
4:      **If** $|(NN(x) \cap T_{maj})| > 0$ **and** $|(NN \cap T_{maj})| < k$
5:           $T_{min} = T_{min} \cup \{x\}$
6:      End **if**
7: End **for**
8: **for** each $x \in P - T_{min}$ **do**
9:      $NN(x)$: $k$ closest neighbor of x
10:     **If** $|(NN(x) \cap T_{min})| > 0$
11:          $T_{min} = T_{min} \cup \{x\}$
12:     End **if**
13: End **for**
*// majority sample selection*
14: $A \leftarrow$ frequency table of majority samples;
15: **for** each $x \in P$ **do**
16:     $NN(x)$: $k$ closest neighbor of $x$
17:     $NN_{maj}(x)$: majority samples of $NN(x)$
18:     **for** each $y \in NN_{maj}$ **do**
19:          $A_y = A_y + 1$
20:     End **for**
21: End **for**
22: **for** each $x \in N$ **do**
23:     **If** $0 < A_y < k/2$
24:          $T_{maj} = T_{maj} \cup \{x\}$
25:     End **if**
26: End **for**
27: $T_{anch} = T_{min} \cup T_{maj}$
*// extra step to achieve anchor class balance*
**If** $T_{min,anch} < T_{maj,anch}$
$C_{min,anch} \leftarrow$ *candidate minority samples by dynamically increasing nearest neighbors*
*Randomly select from $C_{min,anch}$ until class balance or all minority samples are chosen*
**Else**
$C_{min,anch} \leftarrow$ *candidate minority samples by dynamically increasing nearest neighbors*
*Randomly select from $C_{min,anch}$ until class balance*

---

### 3.1.3 Establish balanced anchor dataset

To regulate the disparity in the minority and majority anchor samples, we perform a technique by incrementing the parameter $k$ to candidate more anchor samples from the underrepresented class. However, modifying $k$ will not impact the former delegated anchor samples. Dynamically raising $k$ will alleviate the sensitivity to the hyperparameter $k$ and increase the adaptability and robustness of the anchor selection scheme. The $k$ is adjusted until anchor candidates surpass the required number for class balance. Finally, instances are randomly selected to achieve the desired balanced anchor data.

## 3.2 Training classifier and sample filtering

Anchor samples will be used to train a multilayer perceptron classifier. We already balanced the ratio between the classes by performing anchor sample selection. Therefore, the classifier will not have a preference towards the dominant class. Furthermore, eliminating the majority class instances in the overlapping zone helps the decision boundary to separate the feature space smoothly. The standalone trained MLP network performs better than a base classifier trained on original imbalance data. Thus, it can transfer primary knowledge about the underlying class distribution and assist the generative model with adequate information.

After the training step, the classifier will filter outliers and hard samples from both classes. Outliers can hinder learning, leading to poor generalization of supervised algorithms. Thus, removing outliers aids the model's capability to distinguish between classes. Complex or hard samples are challenging to classify, and often, classifiers fail to label them correctly. These samples are responsible for the intricate boundary region, which can destabilize the GAN training. Therefore, eliminating them can facilitate the generative model's learning procedure. The MLP classifier will assign labels to the original dataset and anchor samples. Any misclassified sample will be discarded from the data, and the final training data will only consist of all correctly classified samples. They will be used in the Arch-SCGAN training. The training of the MLP classifier and sample filtering process are shown in algorithm 2.

---

**Algorithm 2** classifier training and sample filtering.

---

**Input:** selected anchor set $T_{anch}$; training set $T$; the number of iterations *epoch*; batch size *m*.
**Output:** trained classifier $C$; clean training set $T_{clean}$; clean training set $T_{anch\text{-}clean}$
1: **for** ( i = 1; i ≤ $epoch$; i = i + 1) do
2:     select batch $\{(x^{(i)}, y^{(i)})\}_{i=1}^{m}$ from $T_b$
3:     $\theta^{(C)} \leftarrow adam(\nabla_C L_C)$ update the weights of classifier $C$ using loss function $L_C$
4: End **for**
5: $T_{clean}$ ={} ; $T_{anch\text{-}clean}$ ={} ;
9: **for** each $x, y \in T$ do
10:     $y' \leftarrow$ assign label to x using classifier $C$
11:     **If** y = y'
12:         $T_{clean} = T_{clean} \cup \{x\}$
13:     End **if**
14: End **for**
15: **for** each $x, y \in T_b$ do
16:     $y' \leftarrow$ assign label to x using classifier $C$
17:     **If** y = y'
18:         $T_{anch\text{-}clean} = T_{anch\text{-}clean} \cup \{x\}$
19:     End **if**
20: End **for**

## 3.3 Over-sampling using Anchor stabilized conditional GAN (Anch-SCGAN)

In this section, first we introduce the conditional GAN and illustrate our modifications to it. Then, we outline the overall steps for training and oversampling using the Anch-SCGAN.

### 3.3.1 Conditional generative adversarial network

A generative adversarial network (GAN) is a type of neural network in the category of generative models. It comprises two networks, a generator (G) and a discriminator (D). The generator is supplied with a random vector, denoted as z, which follows a predetermined distribution. The generator then attempts to produce a synthetic sample, defined as G(z), that closely resembles the original samples. Conversely, the discriminator denoted as D(x) is presented with original and fake samples and learns to differentiate the synthetic samples fed by the generator while correctly detecting original data. They compete during training due to the opposing objectives of the two networks. The generator's capacity to mimic original distribution is constantly enhanced, while the discriminator's capacity to distinguish between false and original data is also improved. This process of alternating the training of two networks will continue until they achieve the state of Nash equilibrium. The objective function L(D,G) in Eq. (1) describes the mathematical expression of GAN. In the equation, $P_r$ is the data distribution and $P_z$ represents uniform or a Gaussian distribution and z is sampled from $P_z$:

$$L_{GAN}(D, G) = E_{x \sim p_r}[\log D(x)] + E_{z \sim p_z}\left[\log\left(1 - D(G(z))\right)\right] \tag{1}$$

The discriminator aims to maximize the objective function by making D(x) converges to one and D(G(z)) to zero. For the generator, the aim is to minimize the objective function, which will lead to the value of D(G(z)) converging to one. The generator and discriminator independently optimize parameters in order with their defined objectives. Once the training is over, the generator has the ability to produce samples that nearly reflect real samples.

The variant of GAN that integrates supplementary information is the Conditional Generative Adversarial Network (CGAN). This extra information may represent the class label or any external data provided during training. The architecture and training procedure of CGAN are similar to those of GAN. Eq. (2) introduces the objective function of CGAN. This function is similar to Eq. (1), but it incorporates a condition, denoted as y, by combining it with the input. This condition then influences the outputs of both the generator and discriminator. By including extra information y, it is possible to maintain control over the created data. In this scenario, the Conditional Generative Adversarial Network (CGAN) may be trained to acquire knowledge of several class distributions and produce samples from any class distribution, given that the input variable y belongs to that specific class.

$$L_{GAN}(D, G) = E_{x \sim p_r}[\log D(x)] + E_{z \sim p_z}\left[\log\left(1 - D(G(z))\right)\right] \tag{2}$$

### 3.3.2 The Anch-SCGAN model

This study aims to provide a technique for addressing the imbalanced learning problem by using an oversampling approach built based on the CGAN scheme. We call it Anchor Stabilized CGAN, or Anch-SCGAN, in short. The Anch-SCGAN is trained on binary class imbalanced data $\{(x_i, y_i)\}_{i=1}^n$. The binary class label $y$ indicates the CGAN's external information. For the binary class tabular data, class label $y$ is a one-dimensional vector which can be concatenated with input data $x$ or noise sample $z$ [38]. The concatenated data will be given as input to the discriminator or the generator. A class label is an important attribute that illustrates the class type of the sample. Representing the class label with a one-dimensional vector attached to the high-dimensional input is a poor representation of the class label. This can affect the discriminator's ability to separate majority and minority class inputs. Moreover, the generator will produce minority samples that are less distinguishable from the majority instances. This can affect the generator's performance in oversampling the minority instances, which is the primary purpose of training CGAN.

To overcome the limitations of the traditional CGAN, we proposed an innovative architecture for the CGAN network. First, the pre-trained classifier's last hidden layer is employed as class representation for the discriminator. A sample is fed into the classifier, and the additional output feature vector is concatenated with the sample data and then fed into the discriminator. The weights of the pre-trained classifier remain fixed through the training process. Furthermore, we introduce the minority and majority generators to train on minority and majority classes independently. In this case, the generators will no longer require the class condition as an input. For the generators, we specify an additional loss function based on contrastive learning and call it anchor loss. The anchor loss aligns the generated samples with the original data.

Contrastive loss is first introduced by Chopra et al. [46], where a distance metric, i.e. Euclidean distance, is used such that similar samples get in close distance and dissimilar samples get pushed away. The equation for contrastive loss is as follows:

$$L_{contastive}(x_i, x_j) = Dist(x_i, x_j | y_i = y_j) + max\left(0, M - Dist(x_i, x_j | y_i \neq y_j)\right) \quad (3)$$

Where $x_i$ and $x_j$ are samples from the same or different classes, $Dist$ is the distance function determining the distance between samples, and M is a hyperparameter that defines the maximum margin between different class samples. Further, Sohn [47] expands the idea of contrastive representation by introducing the *N*-pair loss function to capture the similarity and dissimilarity of the samples. In *N*-pair objective loss, one sample is selected from the same class, and *N*-1 samples are chosen from the opposing class. The equation for this loss function is as follows:

$$L_{N-pair}(x, x^+, \{x^-\}_{N-1}) = -\log \frac{exp(C(x)^T. C(x^+))}{exp(C(x)^T. C(x^+)) + \sum_{i=1}^{N-1} exp(C(x)^T. C(x^-))} \quad (4)$$

This objective function aims to distinguish existing sample $x$, from opposite class with the assist of information it gathers from $x^+$. We employ *N*-pair loss function for anchor samples and call it anchor loss. In Eq. (4), function C is the output of the last hidden layer of the MLP classifier. $x$ is a sample generated by either a minority or majority generator. $x^+$ is an anchor sample from the same class, and $x^-$ from the opposite class. This loss function perfectly aligns with the choice of two generators. The main problem with Conditional GAN is the inseparability of generated class samples, especially on tabular data where the class label is represented with a single input. In our case, the generators will be trained with contrary objectives, which will help them create separable class samples. We define the loss functions for the minority and majority generators in Eq. (5) and Eq. (6), respectively, where the λ coefficients are set manually:

$$L_{minG} = L_{GAN|y=1}(D, G) + \lambda_1 \times L_{anchor|y, y^+=1, y^-=0}(x, x^+, \{x^-\}_{N-1}) \quad (5)$$

$$L_{majG} = L_{GAN|y=0}(D, G) + \lambda_2 \times L_{anchor|y, y^+=0, y^-=1}(x, x^+, \{x^-\}_{N-1}) \quad (6)$$

To pair a positive sample with the generated sample, we can choose an instance from the same class with the most proximity to the generated sample or randomly pick one. For negative samples, we can include all anchor samples from the opposite class as negative samples. However, finding the nearest positive sample for every generated instance is time-consuming. So, instead, we can perform clustering on minority and majority anchor samples independently and select positive and negative samples based on cluster centroids.

$$clustering(\{x_{anch,min}\}, c) \rightarrow \{centr_{1,min}, centr_{1,min}, \cdots, centr_{c,min}\}$$
$$clustering(\{x_{anch,maj}\}, c) \rightarrow \{centr_{1,maj}, centr_{1,maj}, \cdots, centr_{c,maj}\} \quad (7)$$

For every generated sample, negative samples are cluster centroids from the opposite class, and a positive sample is picked randomly from one of the nearest centroids of the same class. During the training phase, the parameters of the generators and discriminator are changed in a cyclical training process. Initially, the weights of the generators are unchanged, and the discriminator is optimized to minimize its loss function. Subsequently, the discriminator and majority generator weights are fixed, and the minority generator parameters are optimized. At last, the majority generator is trained, while the parameters of the two other networks are fixed. The procedure of training Anch-SCGAN is presented in Algorithm 3.

### 3.3.3 Finetuning Anch-SCGAN and oversampling the minority class

The challenges in training CGAN models arise when we deal with imbalanced data. As mentioned, classification algorithms tend toward the class representing the majority. This issue persists across CGAN and other generative models as well. Since samples from minority classes are scarce, CGAN prioritizes the majority class distribution during training with imbalanced data. Consequently, the majority class tends to dominate the produced samples, diminishing the performance of the CGAN as an oversampling tool. To alleviate this, we train Anch-SCGAN on anchor samples, a modified version of the original dataset that ensures a balanced distribution between classes. By adopting this approach, we improve the model's emphasis on minority samples adjacent to the bordering region, allowing it to provide more reliable instances. This targeted approach significantly bolsters the decision-making efficiency of the classification algorithm.

To finetune Anch-SCGAN, we initialize the models' parameters with the values from the last training. Then, Anch-SCGAN is trained on the anchor data using the procedure shown in Algorithm 3. Once Anch-SCGAN is trained, oversampling is carried out using the minority generator. Algorithm 4 illustrates the complete process of oversampling the minority class. The first step is specifying the quantity of minority samples to be generated. Next, a random vector is drawn from a normal distribution and inputted into the minority generator. This process generates a synthetic minority sample. We iterate this technique until we achieve an evenly distributed dataset.

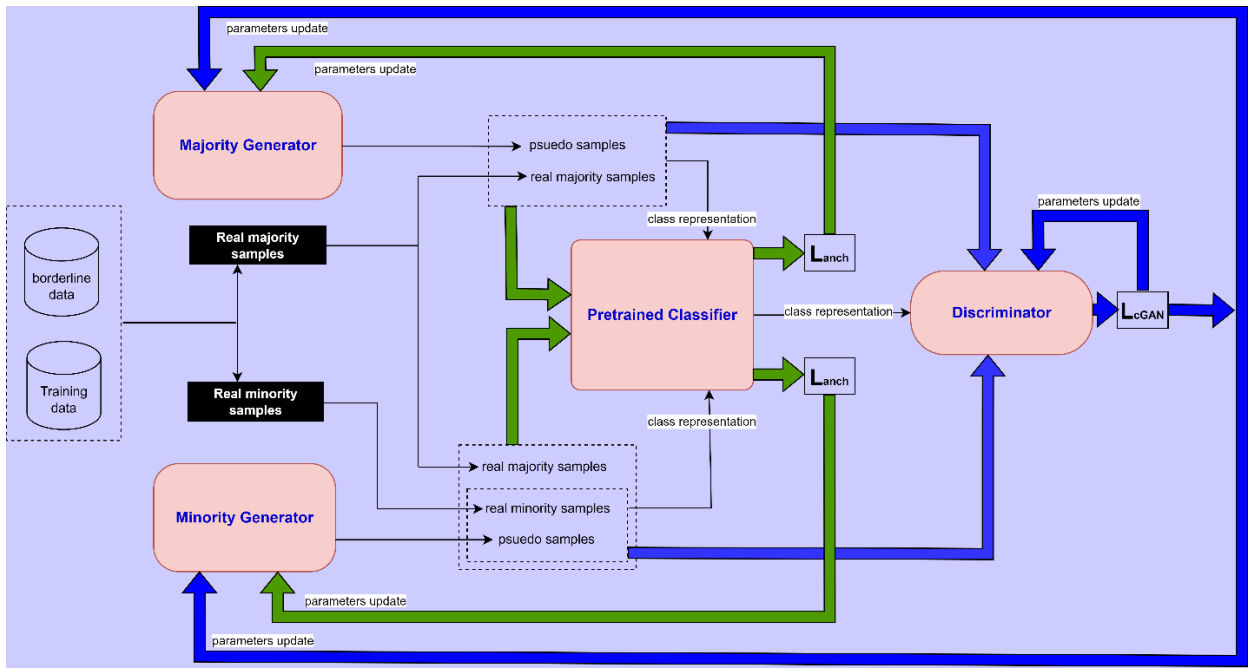

**Fig. 2.** The high-level structure of the Anch-SCGAN model.

**Algorithm 3** Training Anch-SCGAN.

**Input:** clean training set $T$; the number of iterations *epoch*; trained classifier $C$; batch size $m$.
**Output:** model parameters $\theta(D), \theta(G_{min}), \theta(G_{maj})$
1: $T_{min} \leftarrow$ minority instances in $T$; $T_{maj} \leftarrow$ majority instances in $T$
2: **for** ( i = 1; i ≤ *epoch*; i = i + 1) do
// discriminator training
3:       select batch $\{(x^{(i)}, y^{(i)})\}_{i=1}^m \sim P_r$ from $T$
4:       draw batch $\{z^{(i)}\}_{i=1}^m \sim P_z$ from $N\sim(0,1)$
5:       $^{(D)} \leftarrow adam(\nabla_D L_{cGAN})$ update weights of the discriminator using Eq. (1)
// minority generator training
6:       select batch $\{(x^{(i)}, y^{(i)})\}_{i=1}^m$ from $T_{min}$
7:       select batch $\{(x'^{(i)}, y'^{(i)})\}_{i=1}^m$ from $T_{maj}$
8:       draw batch $\{z^{(i)}\}_{i=1}^m$ from $N\sim(0,1)$
9:       $G_{min} \leftarrow adam(\nabla_{G_{min}} L_{minG})$ update weights of the minority generator using Eq. (8)
// majority generator training
10:     select batch $\{(x^{(i)}, y^{(i)})\}_{i=1}^m \sim P_r$ from $T_{maj}$
11:     draw batch $\{z^{(i)}\}_{i=1}^m \sim P_z$ from $N\sim(0,1)$
12:     $G_{maj} \leftarrow adam(\nabla_{G_{maj}} L_{majG})$ update weights of the majority generator using Eq. (9)
13: **End for**
14: **return** $\theta(D), \theta(G_{min}), \theta(G_{maj})$

### 3.3.4 Stabilizing the GAN training

In the training of the GAN network, the generator and the discriminator repeatedly adjust their parameter in response to each other, and due to their adversarial role, they could enter an oscillatory state where the generator could not maintain the quality of generated samples. In this case, even with more epochs, the generator is not guaranteed to achieve better results, or it may even differ from its best performance in earlier epochs. One reason could be that the generator still receives a large gradient from the discriminator despite producing near-real samples [48]. We can use a score function to weigh the generated samples to preserve the generator's ability to generate realistic samples. If the score function assigns a low weight to a generated sample, it is considered a near-real sample, so it will have less influence on the generator's loss function. We will employ the MLP classifier as a score function. In this case, the loss functions for minority and majority generators will change as follows:

$$L_{minG} = (1 - C_o(x)) \times L_{GAN|y=1}(D, G) + \lambda_1 \times L_{anchor|y,y^+=1,y^-=0}(x, x^+, \{x^-\}_{N-1}) \quad (8)$$

$$L_{majG} = C_o(x) \times L_{GAN|y=0}(D, G) + \lambda_2 \times L_{anchor|y,y^+=0,y^-=1}(x, x^+, \{x^-\}_{N-1}) \quad (9)$$

Where $C_o(x)$, the output of the MLP classifier, ranges between zero and one and is trained to outputs near one value for minority class sample and near zero for majority class sample. Suppose the classifier outputs a perfect score for the generated sample. In that case, the weights for the first-term loss will be small, so the generator will have minor changes in its parameters and, thus, be less affected by fluctuation or instability during GAN training.

To test the effect of adding a new coefficient to GAN loss, we utilize a simple toy example called Dirac-GAN, which was introduced by [48]. Dirac-GAN consists of a linear discriminator parametrized by $\psi$, a generator with Dirac distribution parametrized by $\theta$, and real data with Dirac distribution at zero. The Nash equilibrium for Dirac-GAN is at $(\psi, \theta) = (0,0)$. The authors in [48] demonstrate that Dirac-GAN fails to converge to an equilibrium point and instead oscillates in a circle path, as shown in Fig. 3(a). Readers are referred to the main paper for a more in-depth explanation. We introduce a score function $f(x) = e^{-|x|}$, which assigns perfect scores for generated data near real distribution. If we integrate this score into the generator's loss function as a new coefficient, with the proper initial conditions, the training of Dirac-GAN will converge to the optimal point, which is shown in Fig. 3(b) [2].

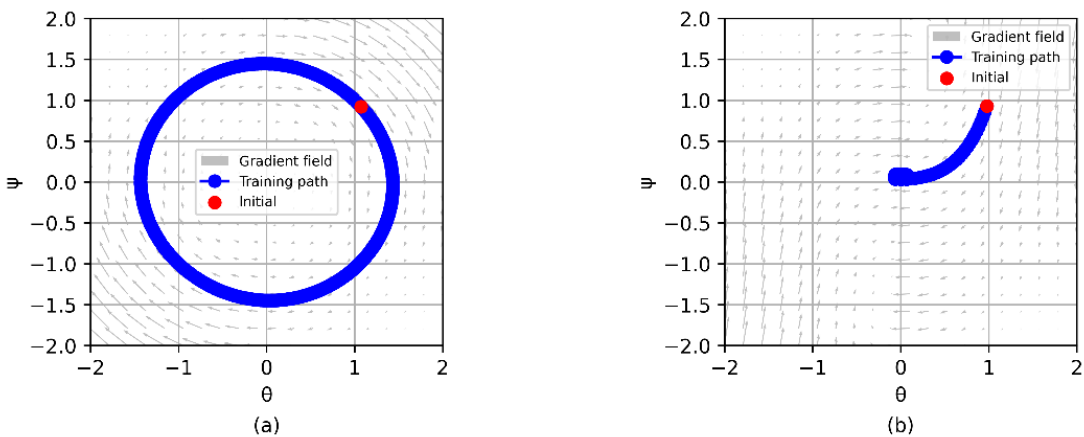

**Fig. 3.** the training path of Dirac-GAN (a) not modifying generator loss (b) modifying generator loss with score coefficient

---

**Algorithm 4** Oversampling with anch-SCGAN.

---

**Input:** training set $T$;
**Output:** balanced dataset $T_{balanced}$
1: $T_{min}$ ← minority instances in $T$; $T_{maj}$ ← majority instances in $T$
2: m = $T_{maj}$ - $T_{min}$;
3: draw Sample $\{z^{(i)}\}_{i=1}^{m}$ from $N \sim (0,1)$
4: $T_{min}$ ← $G_{min}(z)$ generate pseudo samples
5: $T_{balanced}$ = $T_{min} \cup T$
6: return $T_{balanced}$

---

### 3.4 Time complexity analysis

Algorithm 1 outlines the procedure for finding anchor samples. It employs the KNN algorithm and navigates minority samples to find anchor minority and majority samples. With the KD-tree structure, the time complexity for finding nearest neighbors can be reduced to $O(n.d.\log n)$, where n is the total number of samples and d is the sample dimension. Algorithm 2 overviews the MLP classifier's training and sample filtering process. For $T_c$ epochs and $n_{anch}$ anchor samples, the time complexity for training the classifier is $O(T_c.n_{anch})$. An extra data cleaning process is $O(n)$. For the clustering procedure, if we employ k-means clustering for minority and majority anchors, we have $O(n_{anch}.t.c)$ where $c$ is the number of centroids and $t$ is the number of iterations until convergence. The overall computational complexity before Anch-SCGAN training is $O(n.d.\log n + T_c.n_{anch} + n_{anch}.t.c)$. Algorithm 3 reviews the training of Anch-SCGAN. For $T_g$ epochs, the time complexity is $O(T_g \cdot n_{clean} + T_g \cdot n_{gen} \cdot c)$, where $n_{clean}$ is processed dataset, and $n_{gen}$ is generated samples at every epoch. $c$ is the comparison of generated samples with anchor centroids for computing contrastive loss.

---

[2] Code used for plots are from https://github.com/ChristophReich1996/Dirac-GAN/

## 4 Experimental results

The effectiveness of our oversampling scheme is examined through a comprehensive set of experiments. First, we describe the implementation details of the oversampling algorithms, including ours. Next, we introduce the benchmarked dataset and assessment indicators. Finally, we perform extensive analyses of oversampling methods.

### 4.1 Settings

We compare Anch-SCGAN against seven baseline oversampling methods, including four classical oversampling methods: Random Over Sampling (ROS), SMOTE [15], ADASYN [21], Borderline-SMOTE [20], and two GAN-based oversampling methods: CGAN [38], BCGAN [24] and one deep generative model, ConvGeN [45]. We use the default configurations of the baseline methods. The value of $k$ in the KNN search for SMOTE, Borderline-SMOTE, and ASASYN is set to five. In the Borderline-SMOTE method, the value of M, which judges the potential danger of a minority instance, is set to 10, and the procedure used is 'borderline-1'. Moreover, during the training process, we deployed the authors' code for ConvGeN with its default settings.

For the sake of fair comparisons, the hyperparameters and structure of GAN are the same in three GAN-based approaches, including our proposed method. We determine the final configuration based on grid search. Dropout or batch normalization layer is not signed to the discriminator and generator. The network configuration used for Anch-SCGAN is detailed in Table 2. The noise vectors $z$ in the inputs of the generators are set to a size of 100. The number of batches at each epoch is set to 20. In our proposed method, for the KNN search in anchor sample selection, $k$ is set to 5. The sum of epochs in training and finetuning phase is 1,000. The weight coefficients of contrastive loss are set to 0.5. The optimizer is Adam, with the learning rate of 0.001. Moreover, the learning rate of the finetuning stage is set to 0.0005. We also implement the Adam optimizer with its default settings for the MLP classifier. The hyperparameters and the network structure of the MLP classifier are shown in Table 3.

Finally, after performing oversampling, we choose SVM for the classification task, with the default setting for the regularization parameter and the kernel function. We assign 30% of the data as the test set, while the remaining 70% is used as the training set and undergoes oversampling. To mitigate uncertainty in our experiments, we conducted each experiment five times, randomly dividing the dataset and reporting the average results.

### 4.2 Benchmark datasets

We conduct an evaluation of the algorithms across 16 imbalanced datasets from different domains, sourced from the UCI Machine Learning Repository[3] and the KEEL open-source software[4]. The datasets utilized encompass both binary classification (such as Wisconsin and Surgery datasets) and multi-classification datasets (including Ecoli, Yeast, among others). Given our paper's emphasis on binary classification within imbalanced learning contexts, we convert multi-class datasets into a binary format via one-versus-all (OvR) strategy [49]. This involves designating a single class as the minority class and grouping the rest as the majority class. For example, there are 10 class distributions in Yeast (such as NUC, MIT, ME3, EXC). A single class is chosen to serve as the minority class. The majority class is formed by merging the remaining classes. For the datasets employed, the imbalance rate (IR) varies from 1.86 to 85.88, the number of features spans from 7 to 28, and the sample count fluctuates between 214 and 20 000. Table 4 outlines the dataset's properties, including dimensions and class sample proportion.

---

[3] https://archive.ics.uci.edu/datasets
[4] http://www.keel.es/

**Table 2**
The structure setting of Anch-SCGAN.

| Number | generator | setting | Number | discriminator | setting |
|---|---|---|---|---|---|
| 1 | ReLU | (100, 512) | 1 | ReLU | (input, 512) |
| 2 | ReLU | (512, 128) | 2 | ReLU | (512, 128) |
| 3 | ReLU | (128, 32) | 3 | ReLU | (128, 32) |
| 4 | sigmoid | (32, input) | 4 | sigmoid | (32, 1) |

Note: 'input' is the number of attributes in every sample.

**Table 3**
The setting of the MLP classifier.

| Number | Structure | setting | Number | hyperparameter | value |
|---|---|---|---|---|---|
| 1 | ReLU | (input, input ×2) | 1 | epochs | 1,000 |
| 2 | ReLU | (input ×2, input ×4) | 2 | batch size | 8-128 |
| 3 | sigmoid | (input ×4, input) | | | |
| 4 | sigmoid | (input, 1) | | | |

Note: the batch size varies based on the sample size

**Table 4**
detail of the used datasets.

| Dataset | Size | attributes | IR | Min | Maj | Class(majority/minority) |
|---|---|---|---|---|---|---|
| Ecoli2 | 336 | 7 | 5.46 | 52 | 284 | Others/pp |
| Ecoli3 | 336 | 7 | 8.6 | 35 | 301 | Others/imU |
| Ecoli4 | 336 | 7 | 15.8 | 20 | 316 | Others/om |
| Wine-red789 | 1599 | 11 | 6.4 | 217 | 1382 | Others/over 7 |
| Surgery | 470 | 24 | 5.71 | 70 | 400 | False/True |
| Letter | 20 000 | 10 | 25.05 | 768 | 19 232 | Others/letter E |
| CMC2 | 1473 | 21 | 3.42 | 333 | 1140 | Others/2 |
| Wisconsin | 683 | 9 | 1.86 | 239 | 444 | Benign/Malignant |
| Yeast1 | 1484 | 8 | 2.46 | 429 | 1055 | Others/NUC |
| Yeast2 | 1484 | 8 | 5.1 | 244 | 1240 | Others/MIT |
| Yeast3 | 1484 | 8 | 8.1 | 163 | 1321 | Others/ME3 |
| Yeast6 | 1484 | 8 | 41.4 | 35 | 1447 | Others/EXC |
| Glass2 | 214 | 9 | 11.59 | 17 | 197 | Others/vehicle windows |
| Vowel0 | 988 | 13 | 9.98 | 90 | 989 | Others/0 |
| Poker86 | 1477 | 10 | 85.88 | 17 | 1460 | 8/6 |
| Thyroid | 3772 | 28 | 64.03 | 58 | 3714 | Negative/ Discordant |

### 4.3 Evaluation metrics

When studying imbalanced learning, relying solely on 'accuracy' to evaluate a classifier's effectiveness is inadequate. This is because accuracy predominantly measures the overall performance of the learning algorithm, rather than providing insight into the classifier's precision for the minority class. Therefore, we employ a set of evaluation metrics designed to yield a more insightful experimental assessment in the frame of imbalanced learning.

Three widely used evaluation metrics in an imbalanced learning framework are the F1-score, geometric mean (G-mean), and the area under the curve (AUC) of the receiver operating characteristic (ROC). The F1-score employs accuracy and recall to indicate the model's performance in correctly classifying minority classes. The metric is formulated through the following equations:

$$precision = \frac{TP}{TP + FP} \tag{10}$$

$$recall = sensitvity = \frac{TP}{TP + FN} \tag{11}$$

$$F1\_score = \frac{2 \times precision \times recall}{precision \times recall} \tag{12}$$

TP and FN indicate correctly and incorrectly labeled minority class samples, respectively. TN and FP represent correctly and incorrectly labeled instances of the majority class, respectively. The G-mean metric assesses the accuracy of how minority and majority class samples are classified. The formulas for calculating G-mean are as follows:

$$specificity = \frac{TN}{TN + FP} \tag{13}$$

$$G\_mean = \sqrt{recall \times specificity} \tag{14}$$

The AUC, an essential metric obtained from the ROC curve, is also derived by the following formula [27]:

$$AUC = \frac{recall + specificity}{2} \tag{15}$$

For all evaluation metrics, a higher value indicates a better classification outcome.

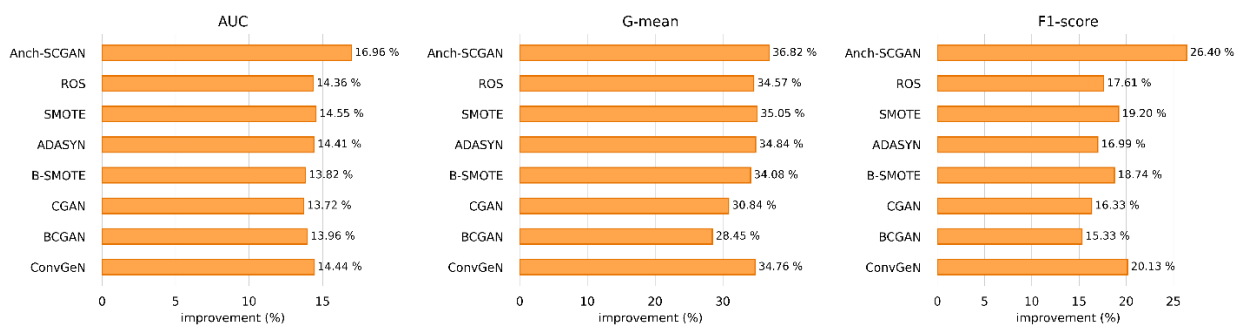

**Fig. 4.** Improvements of various methods compared to the 'original' method in three evaluation metrics: AUC, G-mean and F1-score.

### 4.4 Results

Tables 5, 6 and 7 demonstrate the experimental results for each dataset, delineated according to the respective evaluation metrics. Bolded and underlined numbers are the best and worst oversampling schemes for each dataset. In the tables, the 'original' column denotes the classification performance metrics obtained from the original datasets without oversampling. The AUC and G-mean evaluations are detailed in Tables 5 and 6. According to the AUC results,

proposed model outperforms other oversampling methods: Anch-SCGAN achieves the best results on 11 out of the 16 datasets evaluated and never achieves the worst result on any dataset. Moreover, over the Wine-red789 and Ecoli2 datasets, Anch-SCGAN achieves the second and third-best scores, respectively. Over the Letter dataset, it outperforms the other two approaches based on GANs. ROS, CGAN, BCGAN and ConvGeN score best over two datasets. Fig. 4 depicts the average improvement in classification performance realized by each oversampling technique across all datasets compared to the 'original.' Oversampling methods ROS, SMOTE, ADASYN, B-SMOTE, CGAN, BCGAN and ConvGeN outperform 'original' respectively by 14.36%, 14.55%, 14.41%, 13.82%, 13.72%, 13.96% and 14.44%. Compared to the 'original,' the AUC of Anch-SCGAN exhibits an increase of 16.96%. In comparison to ROS, SMOTE, ADASYN, B-SMOTE, CGAN, BCGAN and ConvGeN, our proposed method demonstrates improvements of 2.6%, 2.41%, 2.55%, 3.14%, 3.24%, 3.00% and 2.52%, respectively.

**Table 5**
AUC scores of the evaluated methods over the 16 datasets.

| Datasets | Original | ROS | SMOTE | ADASYN | B-SMOTE | CGAN | BCGAN | ConvGeN | Anch-SCGAN |
|---|---|---|---|---|---|---|---|---|---|
| Ecoli2 | 0.8933 | 0.9268 | 0.9304 | 0.9117 | 0.9293 | **0.9419** | 0.9373 | 0.9393 | 0.9390 |
| Ecoli3 | 0.7255 | 0.8798 | 0.8814 | 0.8737 | 0.8784 | 0.9156 | 0.9188 | 0.8712 | **0.9356** |
| Ecoli4 | 0.8667 | 0.9034 | 0.8982 | 0.9085 | 0.9052 | 0.9513 | 0.9554 | 0.9114 | **0.9675** |
| Wine-red789 | 0.6054 | **0.8156** | 0.7963 | 0.8019 | 0.7959 | 0.7927 | 0.7914 | 0.8123 | 0.8139 |
| Surgery | 0.5000 | 0.5690 | 0.5535 | 0.5577 | 0.5471 | 0.5000 | 0.5000 | 0.4905 | **0.5916** |
| Letter | 0.6433 | **0.9764** | 0.9743 | 0.9705 | 0.9584 | 0.8841 | 0.9036 | 0.9743 | 0.9558 |
| CMC2 | 0.5000 | 0.6623 | 0.6632 | 0.6579 | 0.6649 | 0.6697 | 0.6641 | 0.6697 | **0.6807** |
| Wisconsin | 0.9653 | 0.9699 | 0.9710 | 0.9789 | 0.9785 | 0.9734 | 0.9764 | 0.9766 | **0.9821** |
| Yeast1 | 0.5924 | 0.7042 | 0.7013 | 0.7026 | 0.6988 | 0.7056 | 0.7046 | 0.7138 | **0.7280** |
| Yeast2 | 0.6915 | 0.7791 | 0.7807 | 0.7665 | 0.7563 | 0.7825 | 0.7924 | 0.7753 | **0.8053** |
| Yeast3 | 0.8187 | 0.9225 | 0.9178 | 0.9212 | 0.9194 | 0.9076 | 0.9015 | 0.9162 | **0.9314** |
| Yeast6 | 0.5000 | 0.8972 | 0.9009 | 0.8930 | 0.9017 | 0.9217 | 0.9231 | 0.9282 | **0.9487** |
| Glass2 | 0.5000 | 0.5974 | 0.6094 | 0.6115 | 0.5504 | 0.6303 | 0.6506 | 0.6283 | **0.6661** |
| Vowel0 | 0.9778 | 0.9900 | 0.9899 | 0.9901 | 0.9901 | **1.0000** | **1.0000** | **1.0000** | **1.0000** |
| Poker86 | 0.5400 | 0.8198 | 0.8795 | 0.8800 | 0.8823 | 0.7546 | **0.9331** | 0.8200 | 0.8967 |
| Thyroid | 0.5000 | 0.7029 | 0.6995 | 0.6987 | 0.6739 | 0.6829 | 0.5000 | **0.7031** | 0.6900 |
| Best / Worst | | 2 / 2 | 0 / 2 | 0 / 2 | 0 / 3 | 2 / 2 | 2 / 3 | 2/2 | **11 / 0** |

Meanwhile, the G-mean metrics evaluation results show that Anch-SCGAN achieves the best result on 10 datasets and never achieves the worst result on any dataset. ROS and CGAN achieve the best results on three and two datasets. In addition, ADASYN, BCGAN and ConvGeN achieve the best results on one dataset. Anch-SCGAN achieves the second and third-best results on Wine-red789 and Ecoli3, respectively. As illustrated in Fig. 4, ROS, SMOTE, ADASYN, B-SMOTE, CGAN, BCGAN and ConvGeN outperform 'original' by 34.57%, 35.05%, 34.84%, 34.08%, 30.84%, 28.45% and 34.76%, respectively. In comparison to 'original,' ROS, SMOTE, ADASYN, B-SMOTE, CGAN, BCGAN and ConvGeN, our proposed method shows improvements of 36.82%, 2.25%, 1.77%, 1.98%, 2.74%, 5.98%,

8.37% and 2.06%, respectively. As a result, Anch-SCGAN is capable of enhancing the accuracy of classifying minority and majority classes.

Finally, the F1-score results are presented in Table 7. The F1-score is directly associated with precision and recall, focusing on the model's accuracy in recognizing minority samples. As seen in the table, Anch-SCGAN outperforms the other oversampling techniques by accomplishing the top score on 14 datasets. Moreover, it does not achieve the worst score on any dataset. BCGAN obtains the best scores on Ecoli4 and Vowel0 datasets. Moreover, B-SMOTE, CGAN, and ConvGeN yield the best results for one dataset. Anch-SCGAN achieves the second-best score on Letter. As reported in Fig. 4, the average relative improvements with respect to 'original' across all datasets obtained by ROS, SMOTE, ADASYN, B-SMOTE, CGAN, BCGAN and ConvGeN are respectively 17.61%, 19.20%, 16.99%, 18.74%, 16.33%, 15.33% and 20.13%. Compared to the 'original,' the F1-score of Anch-SCGAN exhibits an increase of 26.40%. Furthermore, Anch-SCGAN outperforms ROS, SMOTE, ADASYN, B-SMOTE, CGAN, BCGAN, and ConvGeN, showing improvements of 8.79%, 7.2%, 9.41%, 7.66%, 10.07%, 11.07%, and 6.27%, respectively. The F1-score results demonstrate that Anch-SCGAN enhances the classification performance of minority samples.

**Table 6**
G-mean scores of the evaluated methods over the 16 datasets.

| Datasets | Original | ROS | SMOTE | ADASYN | B-SMOTE | CGAN | BCGAN | ConvGeN | Anch-SCGAN |
|---|---|---|---|---|---|---|---|---|---|
| Ecoli2 | 0.8891 | 0.9265 | 0.9300 | 0.9113 | 0.9290 | 0.9382 | 0.9364 | 0.9393 | **0.9413** |
| Ecoli3 | 0.6840 | 0.8769 | 0.8789 | 0.8700 | 0.8755 | **0.9350** | 0.9182 | 0.8704 | 0.9149 |
| Ecoli4 | 0.8550 | 0.8979 | 0.8921 | 0.9041 | 0.9004 | 0.9493 | 0.9531 | 0.9081 | **0.9666** |
| Wine-red789 | 0.4725 | **0.8137** | 0.7959 | 0.8011 | 0.7957 | 0.7899 | 0.7912 | 0.8112 | 0.8129 |
| Surgery | 0.0000 | 0.5496 | 0.5499 | **0.5546** | 0.5401 | 0.0000 | 0.0000 | 0.4781 | 0.5224 |
| Letter | 0.5356 | **0.9763** | 0.9743 | 0.9705 | 0.9583 | 0.8792 | 0.9005 | 0.9742 | 0.9556 |
| CMC2 | 0.0000 | 0.6594 | 0.6613 | 0.6497 | 0.6614 | 0.6668 | 0.6543 | 0.6660 | **0.6756** |
| Wisconsin | 0.9653 | 0.9698 | 0.9709 | 0.9787 | 0.9780 | 0.9733 | 0.9763 | 0.9766 | **0.9820** |
| Yeast1 | 0.4709 | 0.7031 | 0.6996 | 0.6895 | 0.6837 | 0.6996 | 0.6988 | 0.7078 | **0.7250** |
| Yeast2 | 0.6272 | 0.7748 | 0.7767 | 0.7652 | 0.7546 | 0.7729 | 0.7831 | 0.7607 | **0.8002** |
| Yeast3 | 0.8004 | 0.9223 | 0.9172 | 0.9212 | 0.9193 | 0.9063 | 0.9001 | 0.9161 | **0.9309** |
| Yeast6 | 0.0000 | 0.8954 | 0.8984 | 0.8912 | 0.8998 | 0.9201 | 0.9209 | 0.9266 | **0.9478** |
| Glass2 | 0.0000 | 0.5946 | 0.6070 | 0.6079 | 0.5469 | 0.6035 | 0.6426 | 0.6283 | **0.6628** |
| Vowel0 | 0.9774 | 0.9899 | 0.9898 | 0.9900 | 0.9900 | **1.0000** | **1.0000** | **1.0000** | **1.0000** |
| Poker86 | 0.1789 | 0.7708 | 0.8638 | 0.8648 | 0.8670 | 0.7427 | 0.9318 | 0.7946 | 0.8904 |
| Thyroid | 0.0000 | **0.6661** | 0.6577 | 0.6606 | 0.6086 | 0.6143 | 0.0000 | 0.6603 | 0.6188 |
| Best / Worst | | 3 / 1 | 0 / 2 | 1 / 4 | 0 / 3 | 2 / 4 | 1 / 3 | 1 / 0 | **10 / 0** |

Altogether, Anch-SCGAN consistently outperforms in all the evaluative criteria. Traditional oversampling methods exhibit results similar to Anch-SCGAN with respect to AUC and G-mean. Nonetheless, their approach of generating synthetic samples based on local minority class patterns may lead to a sub-optimal generalization of the minority class distribution. This issue results in their weak performance in the F1-score, primarily due to a high number of false positives. ConvGeN demonstrates a higher average enhancement in F1-score compared to traditional oversampling methods, yet it exhibits a marginally lower or similar score in the AUC and G-mean metrics. Compared to conventional methods, CGAN and BCGAN obtain lower scores in all assessment metrics. This can be due to challenges in training GANs, which involve finding a desirable balance between the generator and discriminator, fine-tuning the network's architecture, and regulating hyperparameters. The key reasons for the improvements of Anch-SCGAN are as follows:

1. We introduce a procedure for selecting anchor samples from the minority and majority classes, which are crucial for partitioning the feature space. This primary knowledge from the decision boundary is transferred to the generative model via a pre-trained MLP network, which noticeably facilitates and improves our model's capability in simulating minority class distribution.
2. We enhance the generators' training procedure by implementing an auxiliary anchor loss function directed by contrastive learning. Subsequently, we perform a stabilizing strategy to maintain the generators' strength in producing realistic samples.
3. We fine-tune Anch-SCGAN with anchor samples to guide the model in emulating minority class instances near the decision boundary while simultaneously not ignoring the presence of nearby majority samples.

**Table 7**
F1-scores of the evaluated algorithms over the 16 datasets.

| Datasets | Original | ROS | SMOTE | ADASYN | B-SMOTE | CGAN | BCGAN | ConvGeN | Anch-SCGAN |
|---|---|---|---|---|---|---|---|---|---|
| Ecoli2 | 0.8361 | 0.8244 | 0.8388 | 0.7714 | 0.8315 | 0.8793 | 0.8758 | 0.8333 | **0.8873** |
| Ecoli3 | 0.5409 | 0.5902 | 0.6074 | 0.5691 | 0.6065 | 0.6985 | 0.7004 | 0.5556 | **0.7170** |
| Ecoli4 | 0.8436 | 0.7865 | 0.8020 | 0.7705 | 0.7612 | 0.8389 | **0.8614** | 0.8333 | 0.8164 |
| Wine-red789 | 0.3414 | 0.5438 | 0.5279 | 0.5049 | 0.5176 | 0.5490 | 0.5367 | 0.5714 | **0.5952** |
| Surgery | 0.0000 | 0.2791 | 0.2690 | 0.2743 | 0.2616 | 0.0000 | 0.0000 | 0.2078 | **0.2988** |
| Letter | 0.4430 | 0.6826 | 0.7157 | 0.7220 | **0.7311** | 0.6679 | 0.7057 | 0.6696 | 0.7303 |
| CMC2 | 0.0000 | 0.4689 | 0.4700 | 0.4637 | 0.4714 | 0.4805 | 0.4710 | 0.4759 | **0.4953** |
| Wisconsin | 0.9509 | 0.9553 | 0.9562 | 0.9636 | 0.9635 | 0.9585 | 0.9623 | 0.9588 | **0.9685** |
| Yeast1 | 0.3446 | 0.5795 | 0.5762 | 0.5776 | 0.5738 | 0.5806 | 0.5793 | 0.5892 | **0.6095** |
| Yeast2 | 0.5344 | 0.5746 | 0.5751 | 0.5041 | 0.4921 | 0.6143 | 0.6287 | 0.6100 | **0.6361** |
| Yeast3 | 0.7439 | 0.7555 | 0.7717 | 0.7162 | 0.7190 | 0.7694 | 0.7431 | 0.7395 | **0.8084** |
| Yeast6 | 0.0000 | 0.3181 | 0.3566 | 0.2852 | 0.4515 | 0.5096 | 0.5275 | 0.5807 | **0.6354** |
| Glass2 | 0.0000 | 0.1874 | 0.1977 | 0.1965 | 0.1650 | 0.2374 | 0.2367 | 0.2135 | **0.2500** |
| Vowel0 | 0.9771 | 0.9674 | 0.9663 | 0.9681 | 0.9685 | 1.0000 | 1.0000 | 1.0000 | 1.0000 |
| Poker86 | 0.0000 | 0.7214 | 0.8451 | 0.8500 | 0.8513 | 0.1051 | 0.1799 | 0.7700 | **0.8843** |
| Thyroid | 0.0000 | 0.1387 | 0.1519 | 0.1356 | 0.1874 | 0.2784 | 0.0000 | 0.1667 | **0.4474** |
| Best / Worst | | 0 / 1 | 0 / 1 | 0 / 5 | 1 / 4 | 1 / 3 | 2 / 2 | 1 / 1 | **14 / 0** |

## 4.5 Significance test

In this section, we utilize the Friedman test [50] to more accurately determine how the proposed method statistically differs from competing oversampling techniques. Friedman test is a statistical test that uses average ranking as the evaluation indicator. Fig. 5 illustrates the comparative average performance rankings of all the evaluated methods, utilizing the AUC, G-mean, and F1-score as assessment indicators. The results demonstrate a consistent performance of the proposed method across various evaluation metrics.

We carry out a quantitative analysis of the experimental results, assessing the disparities among all the methods via the significance values (*p*-values). A significant difference between the approaches indicates the need to reject the null hypothesis if the p-value is below the predefined significance threshold. In contrast, a p-value above this threshold indicates no statistically meaningful difference between approaches. The significance level is fixed at 0.05, and the outcomes of the Friedman test are shown in Table 8. Based on the *p*-values in the table, a significant difference among the methods is evident. The average rankings and significance tests provide empirical evidence supporting the advantage and effectiveness of our suggested strategy.

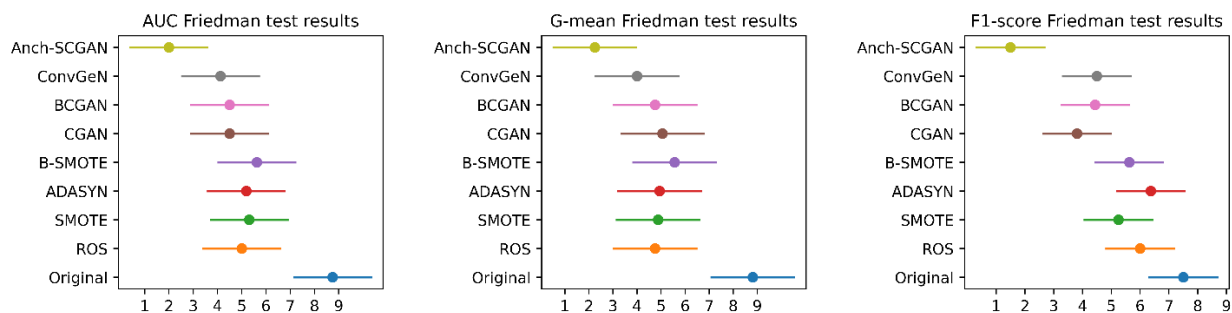

**Fig. 5.** Comparative analysis of average rankings. (a): AUC of average ranking, (b): G-mean of average ranking, (c): F1-score of average ranking.

**Table 8**
Friedman test results.

| Metrics | $p$ | $p < 0.05$ | Hypothesis (H) |
| --- | --- | --- | --- |
| AUC | $6.58 \times e^{-10}$ | True | Rejected |
| G-mean | $2.82 \times e^{-09}$ | True | Rejected |
| F1 | $3.74 \times e^{-09}$ | True | Rejected |

## 4.6 t-SNE analysis

This section illustrates the t-SNE technique for visualizing the data distributions and intuitively perceives the advantages and disadvantages of our sample generation method. t-SNE is a powerful visualization tool that transforms high-dimensional data into more manageable lower dimensions (typically 2D or 3D) that can be visualized. t-SNE preserves the pairwise similarities between points as much as possible. Fig. 6 and 7 visualize the data distribution generated by oversampling methods. The oversampling schemes include four classical methods (ROS, SMOTE, ADASYN, S-SMOTE), GAN-based models (CGAN, BCGAN) and a single deep generative model (ConvGeN), along with our proposed method (Anch-SCGAN). It can be seen from the figures that:

- ROS replicates existing samples and does not produce any synthetic samples. The other classical approaches (SMOTE, ADASYN and B-SMOTE) interpolate between two nearby minority class samples, often resulting in generated samples that cluster around the original data. This diminishes the generalization capability of generated samples and leads to overfitting during training. Furthermore, in the classical oversampling schemes (typically ADASYN and B-SMOTE) and ConvGeN, when minority samples are created close to the decision border, more noise samples are introduced in the overlap area, leading to classification difficulty.
- Two GAN-based methods (CGAN and BCGAN) generate diverse samples and effectively capture the distribution of the original data. However, they often diverge from the minority class distribution duo to mode collapse and training on an imbalanced dataset. Training conditional GANs on datasets with class imbalances has the potential to neglect instances in minority classes and instead prioritize learning the distribution of the majority class. This problem arises because the majority class is often more prevalent in the dataset and, thus, easier for the model to learn from. Therefore, generated minority samples will be pushed toward majority class data, leading to noise in the majority class data.

- Anch-SCGAN includes contrastive loss functions to address challenges in training with imbalanced data. It restricts the produced samples to abide by minority class regions. Therefore, the synthetic samples will closely mirror the original class distribution. Moreover, fine-tuning the model with selected anchor samples contributes to a more balanced training dataset. This guides the generator to create minority samples with a higher impact on shaping the decision boundary while preventing the generation of minority samples that may converge towards the majority class distribution. Finally, unlike other oversampling schemes, our proposed method generates minority samples that are dominant in the regions neighbored by majority-class samples. This indicates that although Anch-SCGAN seeks to generate minority instances near the boundary, it still generates samples that are less likely to be pushed toward the majority-class region.

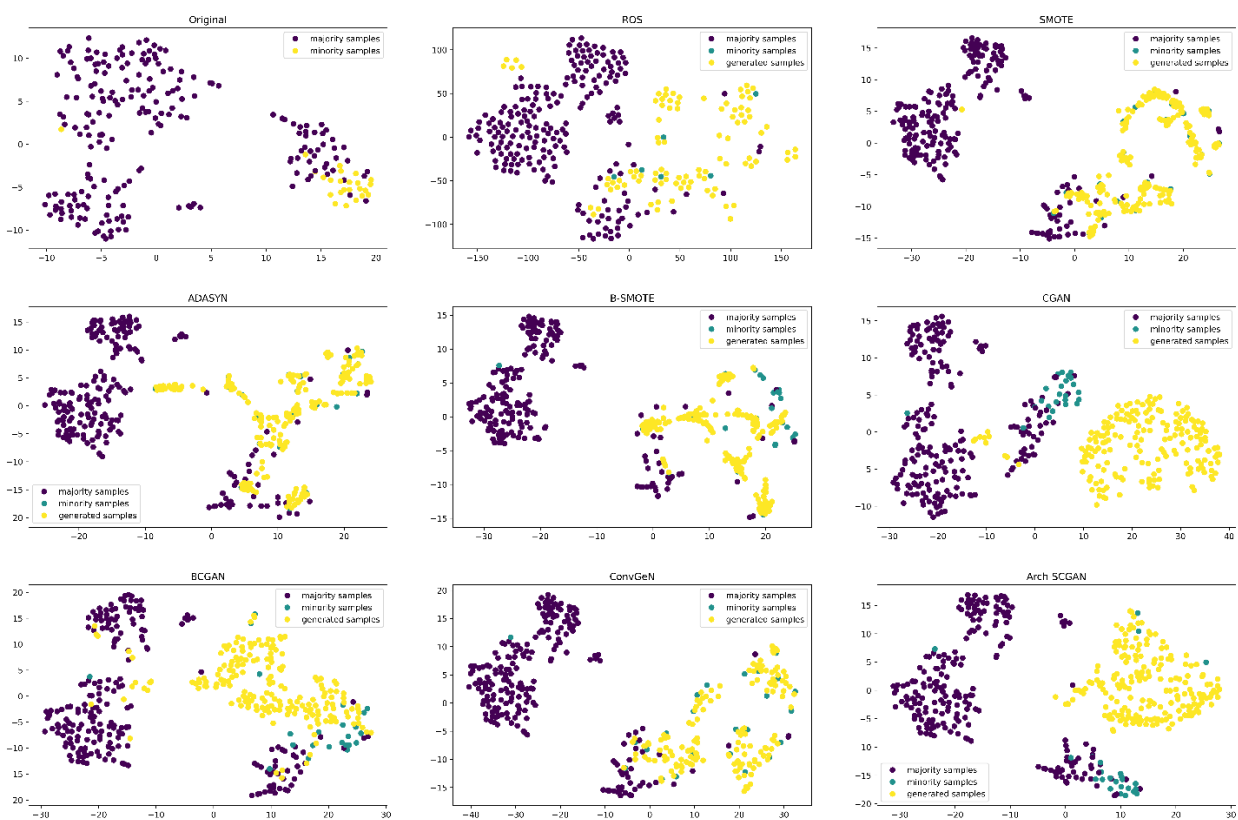

**Fig. 6.** t-SNE visualization in Ecoli3 (purple, blue and yellow samples represent majority class samples, minority class samples and generated minority samples respectively)

### 4.7 Further discussion

**A. anchor selection parameter sensitivity.** The hyperparameter $k$ specifies the proportion of anchor samples used to train the classifier and finetune Anch-SCGAN. Therefore, its value can significantly impact the model's performance. For low values of $k$, only a tiny portion of the minority samples are selected as anchor samples. In the following, the majority anchor samples will be limited by the number of minority anchor samples and cannot exceed a certain amount. In this case, the anchor samples may be less competent in reflecting the decision boundary and providing information about feature space division. The finetuning process could also be less efficient or hazardous in highly imbalanced datasets due to the small size of anchor samples. As *k* increases, the quantity of minority anchor samples will rise, which also affect the proportion of selected majority anchors. For higher *k*, more challenging instances will be labelled as minority anchor samples. Moreover, selected majority anchors will be distant from the decision region because more majority class data will be discarded in the overlapping area. In this case, the authenticity of generated data may degrade as more minority samples will be impelled into the majority class region. Therefore, choosing an appropriate *k* is a key factor in balancing the quality and quantity of anchor samples affecting the performance of Anch-SCGAN. Table 9 shows the experimental results for the Yeast datasets across a range of *k* values. The results show that larger *k* values correlate with reduced F1 scores for datasets with low imbalanced rates. However, a larger *k* value is generally more favorable for highly imbalanced datasets. In this case, more samples from the minority class are selected, which results in better sample generation performance.

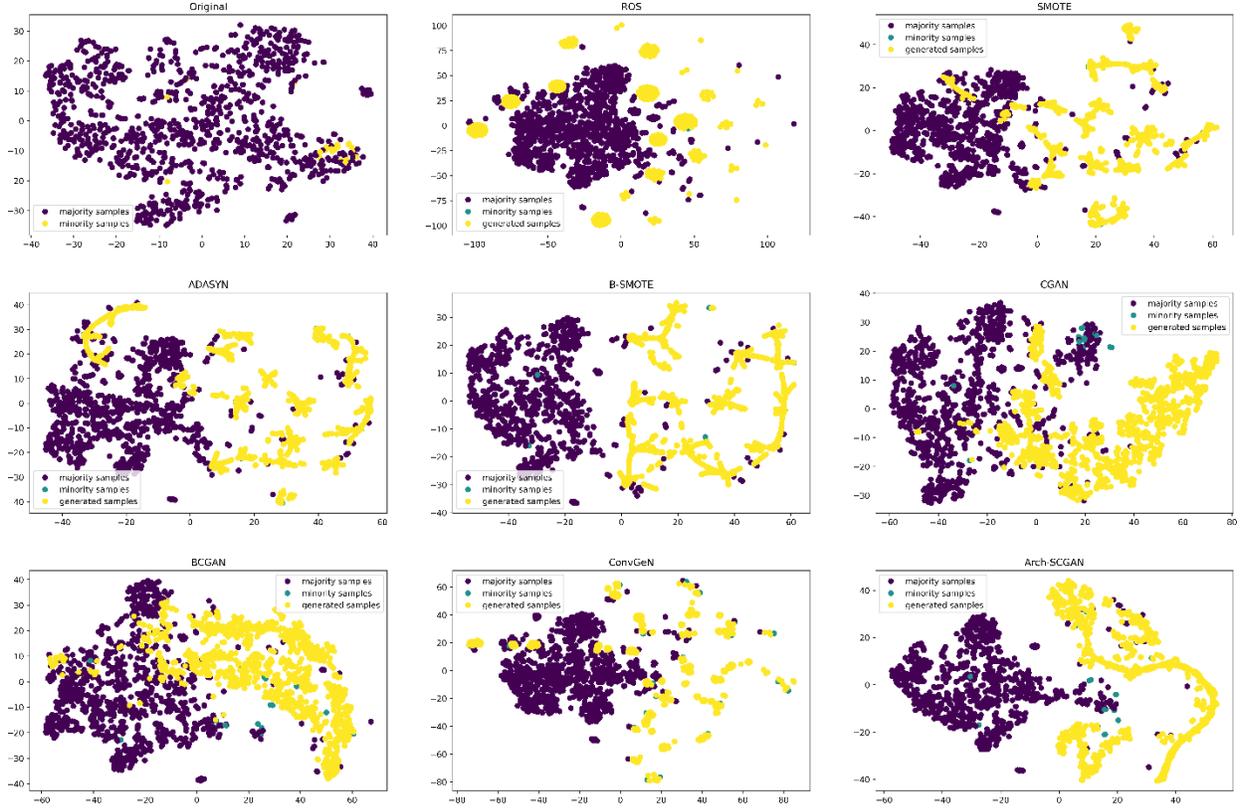

**Fig. 7.** t-SNE visualization in yeast6 (purple, blue and yellow samples represent majority class samples, minority class samples and generated minority samples respectively)

**Table 9**
Results on the Yeast datasets, for three parameters of k.

| Datasets | IR | $k=3$ | | | $k=7$ | | | $k=11$ | | |
|---|---|---|---|---|---|---|---|---|---|---|
| | | AUC | G-mean | F1-score | AUC | G-mean | F1-score | AUC | G-mean | F1-score |
| Yeast1 | 2.46 | **0.7237** | 0.7144 | **0.6114** | 0.7227 | **0.7213** | 0.6042 | 0.7203 | 0.7207 | 0.6026 |
| Yeast2 | 5.1 | 0.7832 | 0.7757 | 0.6018 | **0.7951** | **0.7903** | **0.6277** | 0.7875 | 0.7842 | 0.5907 |
| Yeast3 | 8.1 | 0.9168 | 0.9133 | **0.8143** | 0.9248 | 0.9234 | 0.7896 | **0.9256** | **0.9245** | 0.7756 |
| Yeast6 | 41.8 | 0.8775 | 0.8709 | 0.6334 | 0.9220 | 0.9208 | **0.6516** | **0.9431** | **0.9437** | 0.6405 |

**B. Ablation study.** This part discusses how finetuning affects Anch-SCGAN performance. First, we train the model with the main training set for up to 1000 epochs. Then, we compare the results with those obtained in the previous section. Fig. 8 shows the comparative results of AUC, G-mean and F1-score across 16 datasets before and after finetuning. Over 12 datasets, finetuning has slightly enhanced the final scores. However, improvements in the Ecoli2 and Ecoli3 datasets are too narrow to consider due to the small size of the anchor sample during the finetuning process. In the F1-score metric, finetuning has a notable impact on the Poker86 and Thyroid datasets. Moreover, in the Ecoli4 and Glass2 datasets, which have a higher imbalance ratio and small data size, and the Yeast6 dataset, which has a high imbalance rate, the performance declines after finetuning. Overall, the results suggest that finetuning impacts classification performance.

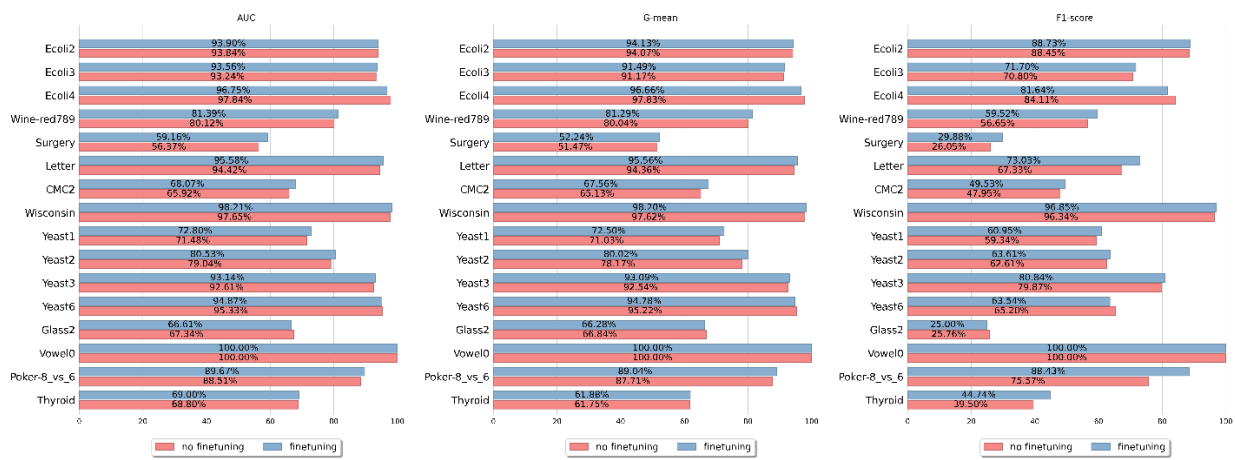

**Fig. 8.** Results of the ablation study (blue bar indicate finetuning stage and red bar indicates no finetuning has performed). (a) AUC results, (b) G-mean results, (c) F1-score results.

## 5 Conclusion and future work

This study presents the Anch-SCGAN model to address the challenge of learning with imbalanced data. Anch-SCGAN first utilizes anchor sample selection, favoring samples in the boundary area. Then, a classifier is trained on boundary anchor samples, which is used to filter misclassified instances from minority and majority classes. Anch-SCGAN employs the final hidden layer of the classifier as the class representation for a conditional GAN model. This study utilizes a novel GAN architecture consisting of two generators, one discriminator and one classifier. In addition, it employs contrastive loss functions to ensure that the generated samples follow the main distribution, thereby enhancing the quality of data generation. In the end, we fine-tuned the trained model on anchor samples to improve the generator's capability to produce minority samples that are important in forming the decision boundary. We empirically evaluated Anch-SCGAN's performance in comparison to four conventional oversampling techniques and two GAN-based approaches. The experimental results across 16 datasets showed that Anch-SCGAN outperforms the comparative methods.

Our proposed model could be further refined by adopting an advanced network architecture that accurately represents classes. This new architecture should impeccably distinguish between classes in the new feature space. Clustering or other preprocessing techniques can also improve the anchor sample selection scheme, another direction for future work. Finally, we can improve the quality of class representation by employing an autoencoder or any suitable architecture for representation learning. It would also be worthwhile to investigate how anchor samples can better participate in training the model and improving the quality of the synthetic samples.